\documentclass[10pt,twocolumn,letterpaper]{article}

\usepackage{wacv}
\usepackage{times}
\usepackage{epsfig}
\usepackage{graphicx}
\usepackage{amsmath}
\usepackage{amssymb}
\usepackage{color}
\usepackage[table]{xcolor}
\usepackage{subfig}
\usepackage{tablefootnote}
\usepackage{url}
\usepackage{sidecap}
\usepackage{xspace}

\def\eg{e.g.\@\xspace } 

\def\ie{i.e.\@\xspace }

\def\cf{c.f.\@\xspace }

\def\vs{vs.\@\xspace }
\def\wrt{w.r.t.\@\xspace } 

\def\etal{et al.\@\xspace }
\def\X{X}
\def\O{O}
\def\L{L}

\def\R{\mathbb{R}}

\wacvfinalcopy
\def\wacvPaperID{****} 

\ifwacvfinal\fi
\begin{document}
	
	\title{Large Scale Novel Object Discovery in 3D}

	
	\author{Siddharth Srivastava\textsuperscript{\textdagger}, Gaurav Sharma\textsuperscript{\textdaggerdbl}, Brejesh Lall\textsuperscript{\textdagger} \\
\textsuperscript{\textdagger} Indian Institute of Technology, Delhi, India, \textsuperscript{\textdaggerdbl}Indian Institute of Technology, Kanpur, India\\
{\tt\small{eez127506@ee.iitd.ac.in, grv@cse.iitk.ac.in, brejesh@ee.iitd.ac.in}}
}

\maketitle
\ifwacvfinal\thispagestyle{empty}\fi
\begin{abstract}
We present a method for discovering never-seen-before objects in 3D point clouds obtained from
sensors like Microsoft Kinect. We generate supervoxels directly from the point cloud data and use
them with a Siamese network, built on a recently proposed 3D convolutional neural network
architecture. We use known objects to train a non-linear embedding of supervoxels, by optimizing the
criteria that supervoxels which fall on the same object should be closer than those which fall on
different objects, in the embedding space. We test on unknown objects, which were not seen during
training, and perform clustering in the learned embedding space of supervoxels to effectively
perform novel object discovery. We validate the method with extensive experiments, quantitatively
showing that it can discover numerous unseen objects while being trained on only a few dense 3D
models. We also show very good qualitative results of object discovery in point cloud data when the
test objects, either specific instances or even categories, were never seen during training.
\end{abstract}

\section{Introduction}

\begin{figure}[t]
	\centering
	\includegraphics[width = 0.5\textwidth]{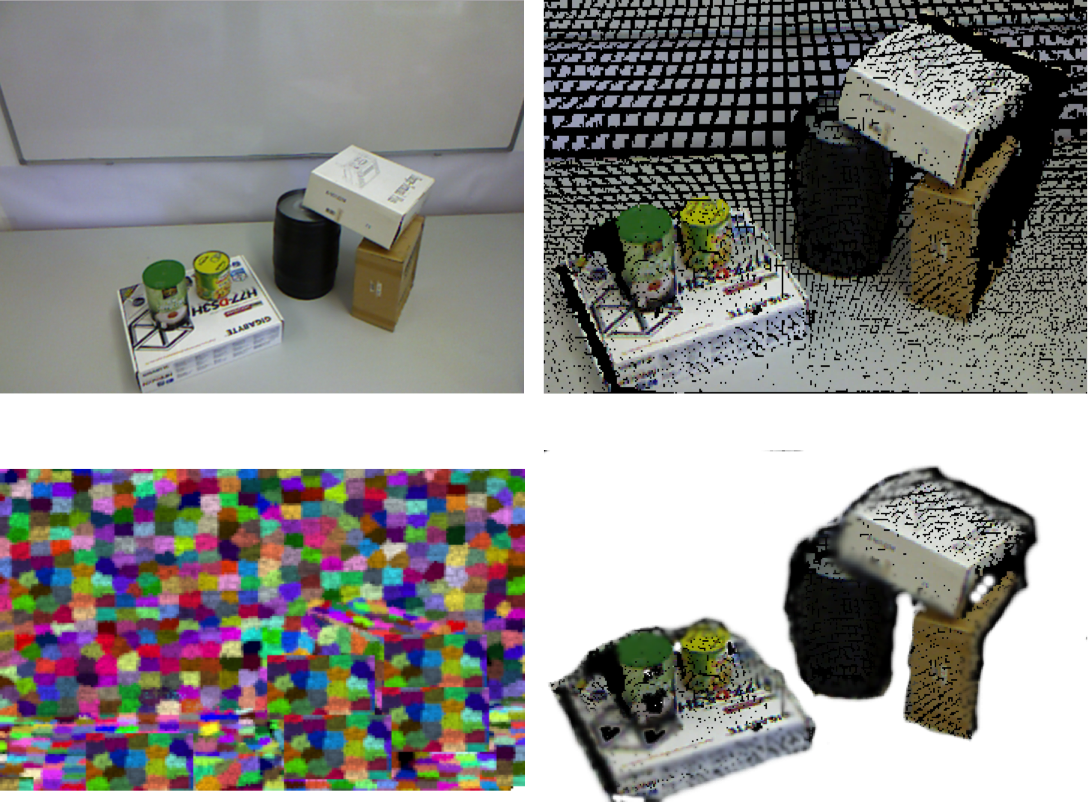}
    \vspace{-2em}
	\caption{
        (Left-top) RGB image of a scene containing multiple objects. (Right-top) Point cloud
        representation of the same scene (noisy). (Left-bottom) Supervoxels obtained from the scene.
        (Right-bottom) Discovered objects (the point cloud has been post-processed for better
        illustration) 
    }
	\label{fig:motivation}
    \vspace{-1em}
\end{figure}

Object discovery has been a popular research area in RGB images \cite{tuytelaars2010unsupervised}
and has recently been explored for applicability on 3D data \cite{karpathy2013object,
mueller2016hierarchical}. As humans perceive the world in 3D \cite{philipona2003there}, the ability
to reason in 3D becomes a necessary computer vision capability. In the case of autonomous mobile
robots, for instance, an important problem is the efficient navigation of the robot in the real world
scenario, where inferring and demarcating manipulable objects and their context is one of the main
technologies required.  Traditionally, such object estimation problem has mostly been well studied
in a supervised \emph{known object} scenario \cite{gupta2015indoor, gupta2014learning}, \ie the
robot is expected to encounter the same objects it was trained on. However, a more realistic and
challenging setting is when the objects that the robot will encounter were never seen before -- we
address this important problem of discovering never before seen objects from 3D data. 

We propose a learning based approach to discover novel objects in 3D. The proposed method (i) deals
with large scale data, \eg the number of points processed in the 3D datasets used are of $\O(10^6)$,
(ii) learns models which have a large number of parameters $\O(10^5)$ and (iii) discovers objects
from approximately $300$ unseen object categories. While processing such large scale 3D data,
efficiency is a key concern, and the proposed method is successfully able to cope with such scale of
both data and models. In our experiments, we show that on a large dataset (with respect to number of
classes, models and number of points), earlier techniques lag behind the proposed method. Moreover, with experiments on multiple datasets, we show that the model learned with
the proposed method on one (large) dataset, provides state-of-the-art results on other datasets with
a standard forward pass on the test point clouds i.e.\ without need for retraining the
network.

We do bottom up aggregation of supervoxels obtained from 3D scenes into objects. We work in the
setting where at test time, on the field, the robot would encounter novel objects which were never
seen while training. The novel perspective comes from the use of discriminative metric learning
approach for obtaining the similarity between supervoxels which are expected to belong to the same
object, which might itself be unknown. There have also been recent attempts to find 3D bounding
boxes around objects, \eg generic object proposals in 3D \cite{chen20153d, DeepSlidingShapes}. Here,
we aim at discovering strict object boundaries in 3D, and not just bounding boxes, which could help,
\eg, in better grasping by a robot. Borrowing from the recent success of deep learning based methods
on (i) 3D data \cite{bu2014learning, eitel2015multimodal, maturana2015voxnet, wang2015large, wu20153d}, (ii) non-linear embeddings, on various computer vision problems, \eg face
verification \cite{schroff2015facenet}, face retrieval \cite{BhattaraiCVPR2016}, semantic image
retrieval \cite{SharmaICCV2015}, and (iii) unsupervised feature learning
\cite{huang2016unsupervised, wang2015unsupervised}, we propose to use a deep Siamese network for the
task of learning non-linear embeddings of supervoxels into a Euclidean space, such that the $\ell_2$
distances in the embedding space reflect object based similarities.

In summary, we address the problem of 3D object discovery in a supervised, but constrained setting
where we are given a few known objects at train time, while, at test time we are expected to
discover novel unknown objects. We train a 3D-CNN using voxelized CAD models, and use supervoxels at
multiple resolutions to learn a distance metric among supervoxels that are similar (same object) and
distinct (different objects). The contributions of this paper are as follows. (i) We propose a
Siamese network based embedding learning framework for supervoxels which works directly with point
cloud data and can be trained end-to-end, (ii) the proposed method is fast, requiring three
sub-steps of  (a) supervoxel computation, (b) a CNN forward pass and (c) classification of
supervoxels with learned discriminative embeddings, (iii) we show with quantitative and qualitative
results, on challenging public datasets of 3D point clouds, that the structure of objects in terms
of distances between supervoxels can be learned from only a few reference models and can be used to
generalize to (supervoxels of) unseen objects, (iv) while previous object discovery methods were
evaluated on a small number of distinct objects ($2\sim10$), we show that the proposed
technique can efficiently detect objects from a large number of classes and object instances per
class ($\sim100$ and $\sim200$ respectively), (v) we show improvements \wrt challenging baselines and
report state-of-the-art results on three public benchmarks, namely NYU Depth v2, Object Discovery Dataset, and Object Segmentation Dataset, on the task of
novel object discovery in 3D point cloud data.

\section{Related Work}

\begin{figure}
	\centering
	\includegraphics[width=\columnwidth,trim=0 410 260 0,clip]{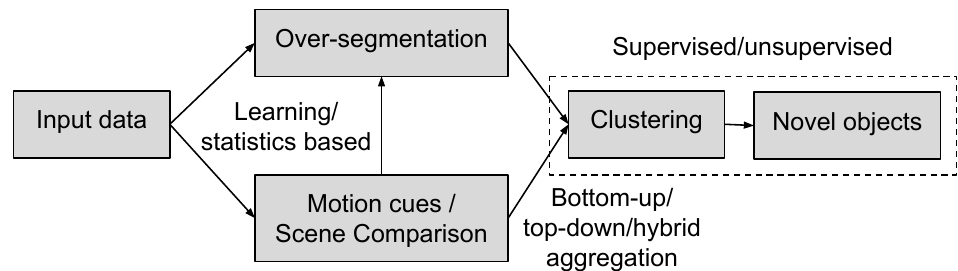}
	\caption{General Pipeline of Object Discovery Methods}
	\label{fig:pipeline}
	\vspace{-1em}
\end{figure}

The general pipeline followed by the majority of the works related to object discovery in 3D is
shown in Fig.~\ref{fig:pipeline}. Among the representative works, Karpathy \etal
\cite{karpathy2013object} discover objects by using multiple views from a 3D mesh. Garcia \etal
\cite{garcia2015saliency} generate object candidates by over-segmenting the RGB-D images. They then fuse the segments, obtained using Voxel Cloud Connectivity Segmentation (VCCS) \cite{papon2013voxel}, with various ranking criterion, to obtain
discovered objects. Compared to these works, we apply discriminative metric learning to learn which
segments should be fused. Meuller \etal \cite{mueller2016hierarchical} first over-segment the RGB-D
point cloud and then learn a classifier for primitive objects. The approach is similar to ours but in our method, we do not
explicitly assume any structural composition of basic shapes that constitute an object, allowing us
to deal with more complex and novel object categories. Richtsfeld \etal
\cite{richtsfeld2012segmentation} perform object discovery on the constrained domain of table-top
scenes. 
Firman \etal
\cite{firman2013learning} learn similarity among segments of an image by training a classifier as
well, while our features are learned using a deep convolutional network trained on CAD models and we learn
the similarity and dissimilarity among segments (supervoxels) with a supervised metric learning based objective.

Authors in \cite{herbst2011toward} propose to discover objects by determining
frequently occurring object instances and determining the patches that move in the scenes.
Shin \etal \cite{shin2010unsupervised} find repetitive objects in 3D point clouds by forming
superpixel like segments of points based on relation between normals of neighbouring points. As a
limitation, they can only find objects which are repeated in a dataset as their method is based on
finding multiple occurring segments. We overcome this limitation by learning the discrimination
measure among supervoxels belonging to same and distinct objects. 

\begin{figure*}
    \centering
	\includegraphics[width=\textwidth, trim=0 400 25 0, clip]{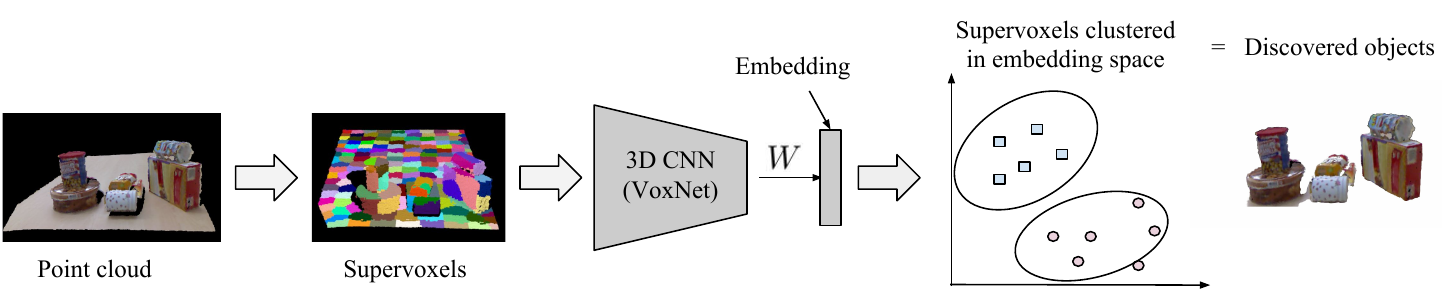}
    \vspace{-2em}
	\caption{Block diagram of the proposed approach at test time. The 3D convolutional network used
    to embed the supervoxels is trained using a Siamese network which captures object based
    similarity of the supervoxels. See Sec.~\ref{secApproach} for details.}
	\label{figBlockdiag}
    \vspace{-1em}
\end{figure*}

Bao \etal \cite{bao2015saliency} perform attention driven segmentation by detecting a salient
object using depth information and then forming a 3D object hypothesis based on the shape
information.  They label and refine objects using Markov Random Field in voxel space with
constraints of spatial connectivity and correlation between color histograms. 
Stein \etal \cite{christoph2014object} decompose a scene into patches while categorizing edges as
convex or concave and thus creating a locally connected sub-graph. Their technique is unsupervised
but mostly detects object parts instead of complete objects. Gupta \etal \cite{gupta2013perceptual}
perform segmentation based on an object's geometric features such as pose and shape. Asif \etal
\cite{asif2016unsupervised} propose perceptual grouping of segments based on geometric features.
Collet \etal \cite{collet2015herbdisc} develop a graph based approach incorporating metadata such as
prior knowledge about objects, environments etc. for discovering objects from raw RGB-D data
streams. While most of the previous methods cater to indoor scenes, Zhang \etal
\cite{zhang2013unsupervised} tackle the problem of discovering category in an urban environment.
They pose object discovery as a problem of determining category structures by finding repetitive
patterns of shape and refining the segmentations by defining an energy function which minimizes
distances between the shape patterns.

\section{Approach}
\label{secApproach}

We now describe our approach for discovering objects in 3D point cloud data.
Fig.~\ref{figBlockdiag} gives the overall block diagram of the approach at test time. At training
time the 3D convolutional neural network is trained in a specific manner which we explain below. We
obtain point cloud data from an appropriate sensor, \eg Microsoft Kinect. The overall strategy is then to discover novel objects in 3D by clustering supervoxels obtained
from the point cloud of a scene. To obtain the distance metric required for clustering, we propose
to use deep Siamese network for learning non-linear embeddings of supervoxels into a Euclidean
space, where the distance reflects the object membership, \ie the supervoxels which belong to the
same object are closer than those which belong to different objects. Learning such a distance
metric requires supervised training data, which we obtain from (the point cloud data of) a set of
few known objects. Note that while the proposed method requires annotated training data, the data it
uses is not from the objects it will see at test time, \ie at test time the method is expected to
discover novel objects which it had not seen during training. Hence, the method is different from the
traditional object detection methods which aim to find the objects, which they were trained on.
Finally, when we have the object based supervoxel embedding, we perform clustering using an
efficient method to obtain object hypothesis. 
We now explain each component in complete detail.

\subsection{Oversegmentation using Supervoxels}
The first step of our method involves obtaining an over-segmentation of the 3D data using
supervoxels. Towards that goal, we use an efficient and robust method to deal with the noisy and
large amount of point cloud data. We group the points in the point clouds into supervoxels using
Voxel Cloud Connectivity Segmentation (VCCS) approach of Papon \etal \cite{papon2013voxel}. Since 
(i) the number of points in the point clouds are very large, $\O(10^6)$, and (ii) they are noisy as
well, this first step of supervoxelization, reduces the computational load on the successive stages
and also lends robustness against noise. Since the objective optimized by the supervoxels creation method
is that they should not cross object boundaries, we aim to thus cluster these supervoxels to give us
plausible novel objects at test time. The seed resolution used to construct supervoxels describes
the fineness (low seed resolution) or coarseness (high seed resolution) of the supervoxel
segmentation, allowing us to have multi-scale supervoxels that can capture fine details as well as
larger blocks of the objects of interest. Hence, we extract supervoxels at multiple seed resolutions
and later merge them appropriately as described below for coherent object hypotheses.  

\subsection{Siamese Deep Network} 
As the next component, we learn a similarity metric which captures object based
distances between the supervoxels, \ie it brings those on the same object closer while pushing those
on different objects far. To do this, we employ a Siamese network which learns a non-linear
embedding into a Euclidean space where the distance captures such semantic relation between the
supervoxels.  

Now, we first outline how we utilize the supervoxels, generated above, with the Siamese network and
then describe the network architecture, loss function as well as training procedure in detail.

Once we have the supervoxels from the point cloud data, we generate `positive' and `negative' pairs
(for each seed resolution) for training the Siamese network as follows. To generate the positive
pairs of supervoxels, we consider all the objects that are labelled during training. For all the
supervoxels $\{z_i | i=1, \ldots, N\}$, in the training set, we create two sets containing all
possible pairs of supervoxels of the following two kinds respectively, (i) the two supervoxels lie
on the same object ($S_+$) or (ii) they lie on different objects or background ($S_-$). To decide if
a supervoxel $z_i$ lies in an object $x \in \X$ (with $\X$ being the set of all objects in the
current scene) or not, we take the intersection of the points $z_i$ with the set of points lying on
the object $x$. If the fraction of these points is more than a threshold $\beta \in \R$, of the total
number of points in the supervoxel, then it is considered to lie on the object. Note that by doing
this we are, effectively, not doing a hard assignment where the supervoxel is required to lie
completely in the object, but doing a soft assignment where the supervoxel is required to be
sufficiently inside the object. Similarly, the set, $S_-$, of negative pairs contains supervoxels
which do not belong to the same object, \ie they either belong to two different objects or to an
object and to the background, respectively.  More details on selecting supervoxels on object and
background for training the network are provided in Sec.~\ref{sec:results}. The following equation
formalizes this mathematically. 
\begin{align}\label{eq:threshold}
    & \forall i, j \in [1, N], \; i \ne j \nonumber \\
	& (z_i, z_j) \in S_+ \;  
    \textrm{ \ iff, } \exists x \in \X \textrm{ s.t. } \nonumber  \\
    & \frac{|pts(z_i) \cap pts(x)|}{|pts(z_i)|} \geq \beta \textrm{ and }
      \frac{|pts(z_j) \cap pts(x)|}{|pts(z_j)|} \geq \beta, \nonumber \\
	& (z_i, z_j) \in S_- \textrm{ otherwise } 
\end{align}
where, $pts(t)$ gives the set of points in the object $t$ and $|\mathcal{A}|$ denotes cardinality of
the set $\mathcal{A}$.

Given such pairs of same and
different object supervoxels, we proceed to train the Siamese network.  Our Siamese network is based
on the recently proposed convolutional neural network for 3D point cloud data by Maturana and
Scherer, called VoxNet \cite{maturana2015voxnet}. VoxNet takes the point cloud as input, computes
the occupancy grid with it and transforms it using two convolutional layers, a max pooling layer and
finally a fully connected layers to finally perform classification over a predefined number of
objects. While it was proposed for object classification in 3D we use it for processing supervoxels
for obtaining their semantic non-linear embeddings.  We have two parallel streams of the VoxNet,
with tied parameters, which take the two members of the positive (or negative) pair and do a forward
pass up to the last fully connected layer (removing the
final classification layer). On top of the last fully connected layer from the two streams, we put a
loss layer based on the hinge loss for pairwise distances given by,
\begin{align}
& \L (\Theta, W) = \sum_{S_+ \cup S_-} \left[ b - y_{ij} (m - \| W f_\Theta (z_i) - W f_\Theta (z_j) \|^2)
\right]_+ \nonumber \\
& \textrm{where, } [a]_+ = \max(a,0) \; \forall a\in R. 
\end{align}
The VoxNet forward pass, parametrized by $\Theta$, is denoted by $f_\Theta(\cdot)$, $W$ is the
parameter of the projection layer after VoxNet, $(z_i,z_j)$ is the pair of supervoxels and
$y_{ij}=+1$ or $-1$ indicates if they are from the same or different objects, \ie if the pair
belongs to $S_+$ or $S_-$, respectively. $b \in \R$ and $m \in \R^+$ are bias and margin
hyper-parameters respectively. 

The loss effectively ensures that in the embedding space obtained after the forward pass by VoxNet
and the projection by $W$, the distances between positive pairs are less than $b-m$, while those
between the negative pairs are greater than $b+m$. When trained on the given train objects, we
expect the loss to capture the semantic distances between the supervoxels in general, which would
then generalize to supervoxels of novel unknown objects at test time. As we show in the empirical
results, we find this to be indeed the case.

\subsection{Supervoxel Clustering and Postprocessing}
Once we have trained the Siamese network, we can embed the supervoxels into a object based semantics
preserving Euclidean space. We then perform clustering in this space to obtain novel object
hypotheses.  We use DBSCAN \cite{ester1996density} to group similar supervoxels into larger
segments. DBSCAN clusters the supervoxels based on density \ie distribution of nearby supervoxel
features. As the features are learned by the deep siamese network to be closer in the embedding
space if they belong to same object, they are considered as potential neighbours by DBSCAN and
vice-versa for those which do not belong to the same object.

We observe that most of the objects were spanned by multiple supervoxels, especially since the
supervoxels were extracted at multiple scales. Therefore, the criteria for a dense region in DBSCAN
was specified as at least two supervoxels in size. This limits the method to discover objects that
consist of at least two supervoxels. In our experiments, we did not find any object with single
supervoxel in the training set, but we do note that in the real world where a robot is navigating, a
single supervoxel may possibly correspond to a heavily occluded object. But it would be important to
note that in case of robot navigation, detecting such heavy occlusions is not a problem since the
the object occluding the view needs to be processed first before moving on to the object behind it.
Moreover, in such cases there are chances that the visible patch of the occluded object becomes a
part of another object. As we show in experiments, the proposed technique is very robust to such
patches and avoids intermixing beyond object boundaries.  

\section{Experimental Results}
\label{sec:results}


\noindent
\textbf{Datasets.} We show results on the following 
challenging and publicly available datasets.

    \textit{NYU Depth v2} \cite{Silberman:ECCV12}: The dataset consists of
    $1449$ RGB-D images with $894$ classes and $35064$ object instances. We use this dataset to
    create a challenging test setting for object discovery. This is in contrast to most of the previous
    works where evaluation was done on much smaller datasets consisting of only a few hundred object
    instances belonging to only a few classes. 
	
    \textit{Universal Training Dataset}: We create a training set from NYU Depth v2 dataset
    with $60\%$ of the classes. We use this as the universal training set for all the experiments
    reported in this paper. 
    We ensure that none of the objects in evaluated test datasets have been seen before. We note
    explicitly here that the \textit{Universal Training Dataset} does not contain any class that are
    present in any of the test data. During testing we only use test split of the corresponding
    datasets and report results on them, except for NYU Depth v2 where we use the remaining classes
    (and instances) for testing. As a scene in NYU Depth v2 consists of multiple classes and
    its instances, any object belonging to the test class is labelled as clutter/background in the
    training set. \footnote{We tried to take only images with the train objects and no test class
    objects but could not create a big enough training set, hence we resorted to the current
    setting. In out training set while some of the test objects may be present, they are not
    annotated (marked as background) and are never used for the supervised training.}
	
    \textit{Object Discovery dataset} \cite{mueller2016hierarchical}: The dataset consists of $30$ RGB-D
    scenes with $640 \times 480$ resolution. There are total $296$ objects where $168$ are
    categorized as simple-shaped and $128$ as complex-shaped. 
	
    \textit{Object Segmentation dataset} \cite{osddataset, richtsfeld2012segmentation}: The dataset
    consists of $111$ scenes categorized into six subsets \ie Boxes, Stacked Boxes, Occluded
    Objects, Cylindrical Objects, Mixed Objects and Complex Scenes. The dataset provides a
    train/test split of $45$ and $66$ point clouds respectively. We used the test set as provided 
    by the authors with the dataset.	

\begin{figure*}[t]
	\centering
	\scalebox{1}{
	\includegraphics[width=0.325\textwidth,trim=20 0 115 0,clip]{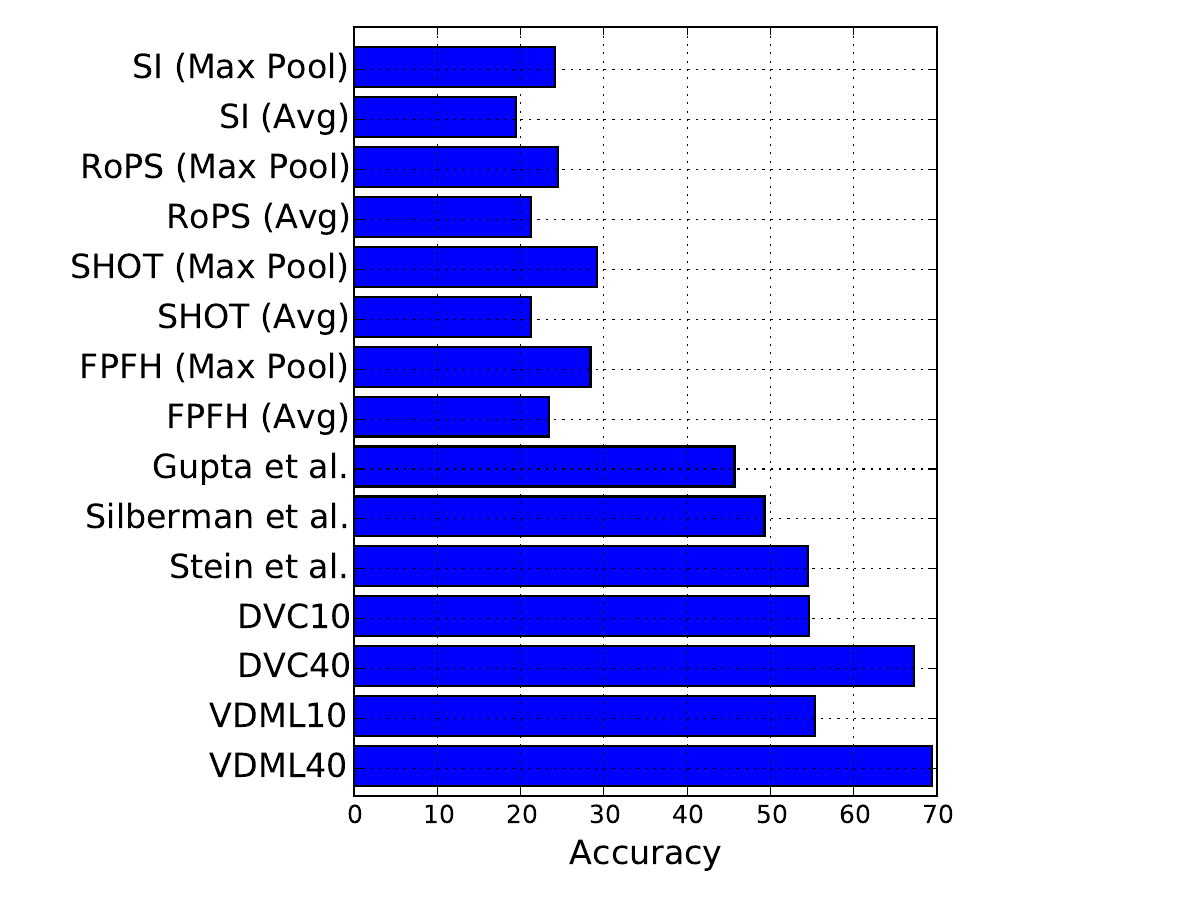} 
	\includegraphics[width=0.325\textwidth,trim=20 0 115 0,clip]{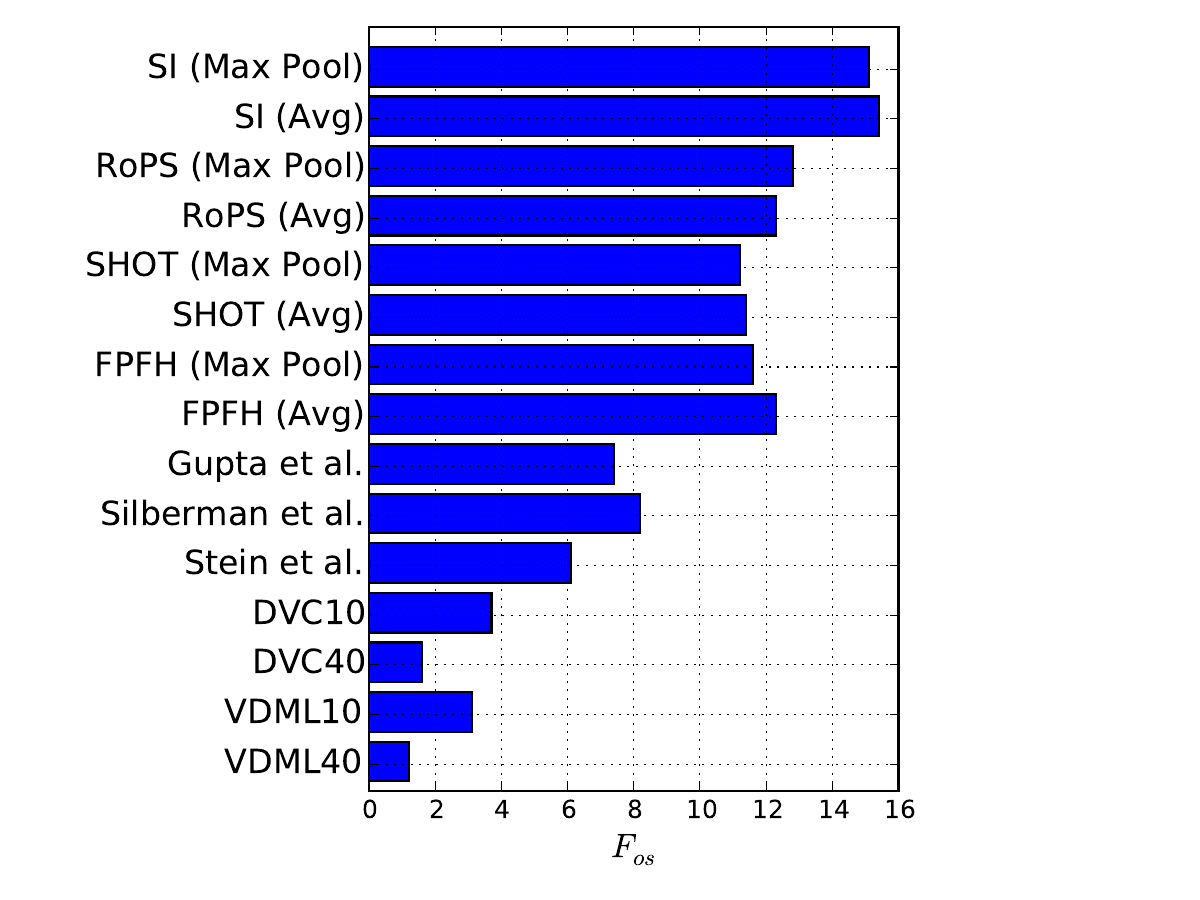}
	\includegraphics[width=0.325\textwidth,trim=20 0 115 0,clip]{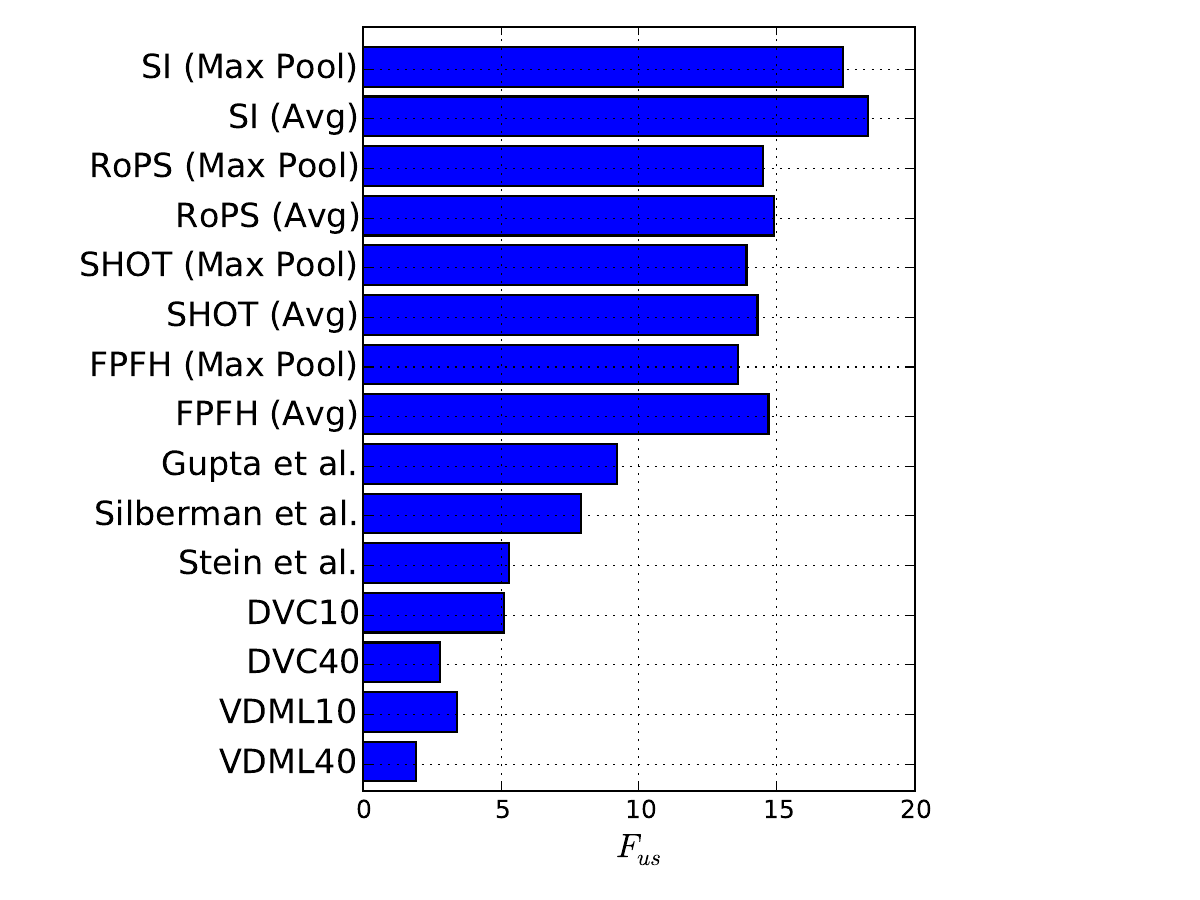}}
    \vspace{-1em}
	\caption{
        Accuracies, $F_{os}$ and $F_{us}$ scores on NYU Depth v2
		dataset. Lower is better for $F_{os}$ and $F_{us}$ scores.
    }
	\label{fig:acc}
    \vspace{-1em}
\end{figure*}

\vspace{0.8em} \noindent
\textbf{Baselines.} We report results with two baselines. The first baseline uses 
keypoint based 3D descriptors while the second is an extended deep architecture based on VoxNet. 


    \textit{3D keypoint based descriptors.} We generate supervoxels from the \textit{Universal
    Training Dataset} and extract various descriptors for points in each supervoxel. The
    cumulative feature descriptor of the supervoxel is calculated by using average pooling and max
    pooling of descriptors. The positive and negative pairs of ground-truth data are prepared using
    the same method as described earlier. We then utilize DBSCAN to find clusters of supervoxels
    which we mark as discovered objects. The evaluated descriptors are Signatures of Histograms of
    Orientations (SHOT) \cite{salti2014shot, tombari2010unique}, Fast Point Feature Histograms
    \cite{rusu2009fast}, Rotational Projection Statistics (RoPS)\cite{guo2013rotational} and Spin
    Image (SI) \cite{johnson1999using} for their superior performance on various metrics for 3D
    computer vision tasks \cite{guo2016comprehensive}. These make for very competitive baselines
    with traditional hand-crafted 3D descriptors.
	
    \textit{Deep VoxNet Classification (DVC) Network}. We remove the last layer of the original VoxNet
    and append two fully connected layers followed by a multiclass classification layer. 
    This method basically classifies each of the training supervoxel into one of the training
    classes. At test time we pass each of the supervoxel through the network and keep the last fully
    connected layer responses as the features of the supervoxels. These supervoxel features are then
    clustered using DBSCAN clustering. This baseline allows us to evaluate the advantages obtained,
    if any, by using the proposed Siamese method.  

\noindent \vspace{-0.2em} \ \\
\textbf{Comparison and evaluation setup.} On NYU Depth v2 dataset, we compare our results against Stein \etal
\cite{christoph2014object}, Silberman \etal \cite{Silberman:ECCV12}, and Gupta \etal
\cite{gupta2013perceptual}. The first one is an unsupervised segmentation technique while the latter
two involve prior training. The technique of Gupta \etal is related to semantic segmentation in 3D,
and is compared here because it reports good performance on NYU Depth v2 and to evaluate if such techniques can
be used for the task of object discovery, \ie when the training and testing set contain mutually
exclusive object categories. Accuracy is computed by counting the number of objects
where the discovered clusters have greater than $80\%$ point-to-point overlap with the ground truth
for classes not in the \textit{Universal Training Dataset}. The over-segmentation ($F_{os}$) and
under-segmentation ($F_{us}$) rates are computed using the method defined in
\cite{richtsfeld2012segmentation} and are given by Eq. \ref{eq:fosfus} where $n_{true}$, $n_{false}$
and $n_{all}$ are number of correctly classified, incorrectly classified and total number of object
points in a point cloud.

\begin{equation}\label{eq:fosfus}
\begin{aligned}
F_{os} = 1 - \frac{n_{true}}{n_{all}}, \       F_{us} = \frac{n_{false}}{n_{all}}
\end{aligned}	
\end{equation}
 
We store a mapping of ground-truth labels of points in point cloud to that of supervoxel and further
in the voxelized cloud. We use these mappings once the supervoxels have been clustered to
obtain $n_{true}$ and $n_{false}$.

On Object Discovery Dataset, the evaluation metrics used are as defined in \cite{mueller2016hierarchical},
which are, over-segmentation ($r_{os}$), under-segmentation ($r_{us}$), good-segmentation ($r_{gs}$)
and mis-segmentation ($r_{ms}$) rates to evaluate various methods on this dataset computed by using labelled segments of object parts. Instead of using object parts, we consider each object as a
segment for comparison on this dataset, which is a more stringent  requirement since the discovered
segments need to correspond to the entire object instead of object parts. The good-segmentation rate is
computed by finding the largest point to point overlap among ground truth and detected objects, while
any clusters partially overlapping with the ground truth objects, contribute to the over-segmentation
rate. The points in ground-truth object which do not belong to any discovered cluster are considered
as mis-segmentations. For remaining discovered clusters, the overlap is calculated with
ground-truth clusters, among those the largest overlap cluster is ignored (as they have already been
considered), while the rest contribute to mis-segmentation rate.  
\vspace{0.8em} \ \\
\textbf{Training details.}
We train the VoxNet using the  implementation corresponding to \cite{maturana2015voxnet} on
ModelNet10 and ModelNet40 \cite{wu20153d} voxelized datasets. We then use this pre-trained VoxNet model with the
proposed Siamese architecture.

Discriminative Metric Learning is performed on top of VoxNet as follows.  We use the
\textit{Universal Training Dataset} described previously for training the Siamese network using
discriminative metric learning. The supervoxels are extracted with seed resolutions of
$0.05$m, $0.10$m, $0.15$m and $0.2$m and subject to the constraint that the \% of ground-truth
object points in the supervoxel should be greater than or equal to a threshold $\beta$. We use the value
of $\beta=0.8$. The value allows for soft-assignment of supervoxels as positive pairs and was chosen
based on empirical evaluation described later (Sec. \ref{subsec:quanteval}). The positive set of supervoxels is constructed
by pairing all supervoxels on the same object for every object in the training data. The negative
set consists of two types of supervoxel pairs.  First are the supervoxels belonging to different
objects, for this we only use supervoxels around the center of the objects to avoid confusion around
boundaries, and second are the pairs of neighbouring supervoxels at object boundaries. We also add
pairs of supervoxels belonging to the background classes to the negative set (wall, floor etc.),
considering them to contribute to hard negatives as they potentially come from very similar objects.
\vspace{0.8em} \ \\
\textbf{Testing details.} 
The test set consists of remaining $40$\% classes of the NYU Depth v2 dataset. The results are
reported on classes in test set only even if some of the test scenes may contain classes from
training set, this may occur since NYU Depth v2 consists of real world complex scenes and having a
clean separation of images with train only and those with test only objects was found to be
infeasible.

\subsection{Quantitative Results} \label{subsec:quanteval}

The results for proposed VoxNet with Discriminative Metric Learning (VDML) on NYU Depth v2 along with the baselines
described above are shown in Fig.~\ref{fig:acc}. VDML outperforms other methods, achieving $55.4$\%
and $69.4$\% test set accuracy on ModelNet10 (VDML10) and ModelNet40 (VDML40) trained 3D-CNN
respectively. It is closely followed by DVC with a test set accuracy of $54.7$\% and $67.3$\% on
ModelNet10 (DVC10) and ModelNet40 (DVC40) respectively. Even training VoxNet with
ModelNet10, we observe that the accuracy is nearly the same as the DVC baseline while being
significantly higher than other baseline methods where the mean performance of descriptors with
average pooling is $21.3$\% while that of max-pooling is $26.6$\%. Another advantage of VDML over
DVC is lower training time. During our experiments, it took $60$\% more time to train DVC as
compared to VDML. This can be explained by the presence of additional fully connected layers in DVC.
Compared to Stein \etal, VDML40 achieves $14.9$\% higher accuracy while method of Silberman \etal
and Gupta \etal lag behind by $20.1$\% and $23.7$\% respectively. This shows that VDML although
following supervised approach is able to generalize better on unseen objects in complex scenes. 

Fig.~\ref{fig:acc} also shows over-segmentation ($F_{os}$) and under-segmentation ($F_{us}$) scores. The
$F_{os}$ for VDML10 and VDML40 are $3.1$\% and $1.2$\% respectively. The corresponding values for
DVC10 and DVC40 are $3.7$\% and $1.6$\% respectively. Comparatively, DVC has higher under-segmentation ($5.1$\% for DVC10 and $2.8$\% for DVC40) than VDML ($3.4$\% for VDML10 and $1.9$\% for VDML40). The low $F_{os}$ and $F_{us}$ demonstrate that
VDML results in lower cross-overs on object boundaries in segmented regions. The other baselines
except DVC perform poorly while methods of Stein \etal, Silberman \etal and
and Gupta \etal perform average with $F_{os}$ being $6.1$\%, $8.2$\% and $7.4$\% respectively. An interesting point to note is the significant reduction in $F_{os}$ and $F_{us}$
scores when compared to training on ModelNet10 and ModelNet40 respectively. It shows that when
the methods are provided with larger number of models, the resulting features are able to better
characterize and learn object boundaries in the embedding space. This underlines the need for more
training data in the 3D domain.


The results on Object Discovery Dataset are shown in Table~\ref{tab:oddresults}. As can be seen in
the results, the proposed method VDML achieves a good-segmentation rate ($r_{os}$) of $98.1$\% on all
objects which is a higher by $5.2$\% when compared to best performing technique (CPC) in
\cite{mueller2016hierarchical}. It would be interesting to note that we obtain significantly reduced
over-segmentation ($r_{os} = 3.2$) and under-segmentation ($r_{us} = 1.7$) rates on all objects
demonstrating that the discovered objects are closer to the complete objects. The mis-segmentation
rate, though, is reduced primarily due to background clutter in a few scenes which are not present as
ground-truth labels in the training set. 

The results on Object Segmentation Dataset are shown in Table~\ref{tab:osdresult}. We achieve a test
set accuracy of $99.24$\% which is $0.83$\% higher than $SVM_{nnb}$ and  $11.2$\%  than $SVM_{nb}$
of \cite{richtsfeld2012segmentation}. Interestingly, we achieve $100$\% accuracy on point clouds
with geometrically simpler Boxes and Stacked Boxes objects, while also having very low, $0$\%
over-segmentation and $0.1$\% under-segmentation rates. These values become more significant when
compared to $SVM_{nnb}$ where higher accuracy is accompanied by significantly large under-segmentation rates \ie $17.2$\% and $28.2$\% as compared to VDML40 with $0.1$\% and $4.4$\% on boxes
and stacked boxes respectively. The $F_{us}$ rate on complex objects for VDML40 is $4.24$\% while
that for $SVM_{nnb}$ and $SVM_{nb}$ are $146$\% and $8.0$\% respectively. $SVM_{nnb}$ witnesses large under-segmentation
rate as it is more sensitive to connection between neighbouring segments. On the
other hand, VDML is trained on multiple seed resolutions of supervoxels allowing it to learn more
intricate connectivity patterns among supervoxel boundaries and hence prevent it from connecting
entire objects if a few segments (supervoxels) are incorrectly assigned to the same object.

\begin{table*}[]
	\centering
	\scalebox{1}{
	\begin{tabular}{|l|r|r|r|r|r|r|r|r|r|r|r|r|r|r|r|}
		\hline
		\textbf{T}        & \multicolumn{5}{c|}{\textbf{Simple-shaped}}                                                                                          & \multicolumn{5}{c|}{\textbf{Complex-shaped}}                                                                                         & \multicolumn{5}{c|}{\textbf{All Objects}}                                                                                            \\ \hline
		\textit{N}        & \multicolumn{5}{c|}{168}                                                                                                             & \multicolumn{5}{c|}{128}                                                                                                             & \multicolumn{5}{c|}{296}                                                                                                             \\ \hline
		\textit{S}        & \multicolumn{1}{l|}{APS} & \multicolumn{1}{l|}{SPS} & \multicolumn{1}{l|}{SPC} & \multicolumn{1}{l|}{CPC} & \multicolumn{1}{l|}{V40} & \multicolumn{1}{l|}{APS} & \multicolumn{1}{l|}{SPS} & \multicolumn{1}{l|}{SPC} & \multicolumn{1}{l|}{CPC} & \multicolumn{1}{l|}{V40} & \multicolumn{1}{l|}{APS} & \multicolumn{1}{l|}{SPS} & \multicolumn{1}{l|}{SPC} & \multicolumn{1}{l|}{CPC} & \multicolumn{1}{l|}{V40} \\ \hline
		\textit{$r_{os}$} & 92.3                     & 43.9                     & 6.6                      & 4.3                      & 2.1                      & 93.1                     & 51.3                     & 22.3                     & 2.8                      & 1.7                      & 92.7                     & 48.8                     & 14.3                     & 4.1                      & 3.2                      \\ \hline
		\textit{$r_{us}$} & 2.4                      & 1.6                      & 1.9                      & 3.7                      & 2.9                      & 1.3                      & 0.9                      & 0.8                      & 0.8                      & 0.5                      & 1.7                      & 1.4                      & 1.6                      & 2.6                      & 1.7                      \\ \hline
		\textit{$r_{gs}$} & 6.8                      & 55.2                     & 92.5                     & 92.3                     & 97.8                     & 6.3                      & 48.1                     & 77.1                     & 94.5                     & 98.3                     & 6.6                      & 50.5                     & 84.9                     & 92.9                     & 98.1                     \\ \hline
		\textit{$r_{ms}$} & 0.8                      & 0.8                      & 0.8                      & 3.4                      & 0.4                      & 0.6                      & 0.6                      & 0.6                      & 2.7                      & 1.6                      & 0.7                      & 0.7                      & 0.7                      & 2.9                      & 1.1                      \\ \hline
	\end{tabular}}

\caption{Comparison on Object Discovery Dataset. Values of APS, SPS, SPC and CPC are from
\cite{mueller2016hierarchical}, V40 refers to VDML40, T = Type of Dataset, N = number of objects, S=
Segmentation Technique and rates $r_{os,us,gs,ms}$ in percentage.}
\label{tab:oddresults}
\end{table*}

\begin{table}[]
	\centering
	\scalebox{0.7}{
	\begin{tabular}{|l|l|l|l|l|l|l|l|l|l|}
		\hline
		\textbf{Technique}  & \multicolumn{3}{c|}{\textbf{SVMnb}}                         & \multicolumn{3}{c|}{\textbf{SVMnnb}}                                            & \multicolumn{3}{c|}{\textbf{VDML40}}                                            \\ \hline
		\textbf{Parameter}  & Acc.  & \multicolumn{1}{c|}{$F_{os}$} & \multicolumn{1}{c|}{$F_{us}$} & \multicolumn{1}{c|}{Acc.} & \multicolumn{1}{c|}{$F_{os}$} & \multicolumn{1}{c|}{$F_{us}$} & \multicolumn{1}{c|}{Acc.} & \multicolumn{1}{c|}{$F_{os}$} & \multicolumn{1}{c|}{$F_{us}$} \\ \hline
		Boxes               & 88.55 & 1.8                      & 0.2                      & 98.19                     & 0.2                      & 17.2                     & 100.0                     & 0.0                      & 0.1                      \\ \hline
		Sta. Box.       & 89.15 & 1.3                      & 7.1                     & 98.99                     & 0.0                      & 28.2                     & 100.0                     & 0.0                      & 4.4                      \\ \hline
		Occ. Obj.    & 87.93 & 16.6                     & 0.1                      & 99.23                     & 0.0                      & 0.2                      & 99.96                     & 0.08                     & 0.1                      \\ \hline
		Cyl. Obj. & 91.66 & 2.6                      & 0.3                      & 96.77                     & 2.6                      & 3.5                      & 99.3                      & 0.7                      & 0.9                      \\ \hline
		Mix. Obj.       & 91.04 & 1.9                      & 19.7                     & 94.97                     & 1.3                      & 1.3                      & 98.5                      & 0.6                      & 1.7                      \\ \hline
		Com. Obj.     & 84.61 & 7.0                      & 8.0                      & 98.97                     & 5.4                      & 146                      & 97.7                      & 2.3                      & 4.2                      \\ \hline
		Total               & 87.72 & 4.5                      & 7.9                      & 98.41                     & 2.7                      & 69.5                     & \textbf{99.24}            & \textbf{0.61}            & \textbf{1.9}                      \\ \hline
	\end{tabular}}
\caption{Comparison with \cite{richtsfeld2012segmentation} on Object Segmentation Dataset. }
\label{tab:osdresult}
\end{table}

\begin{figure}[t]
	\centering
	\includegraphics[width=0.9\columnwidth,trim=0 50 0 80, clip]{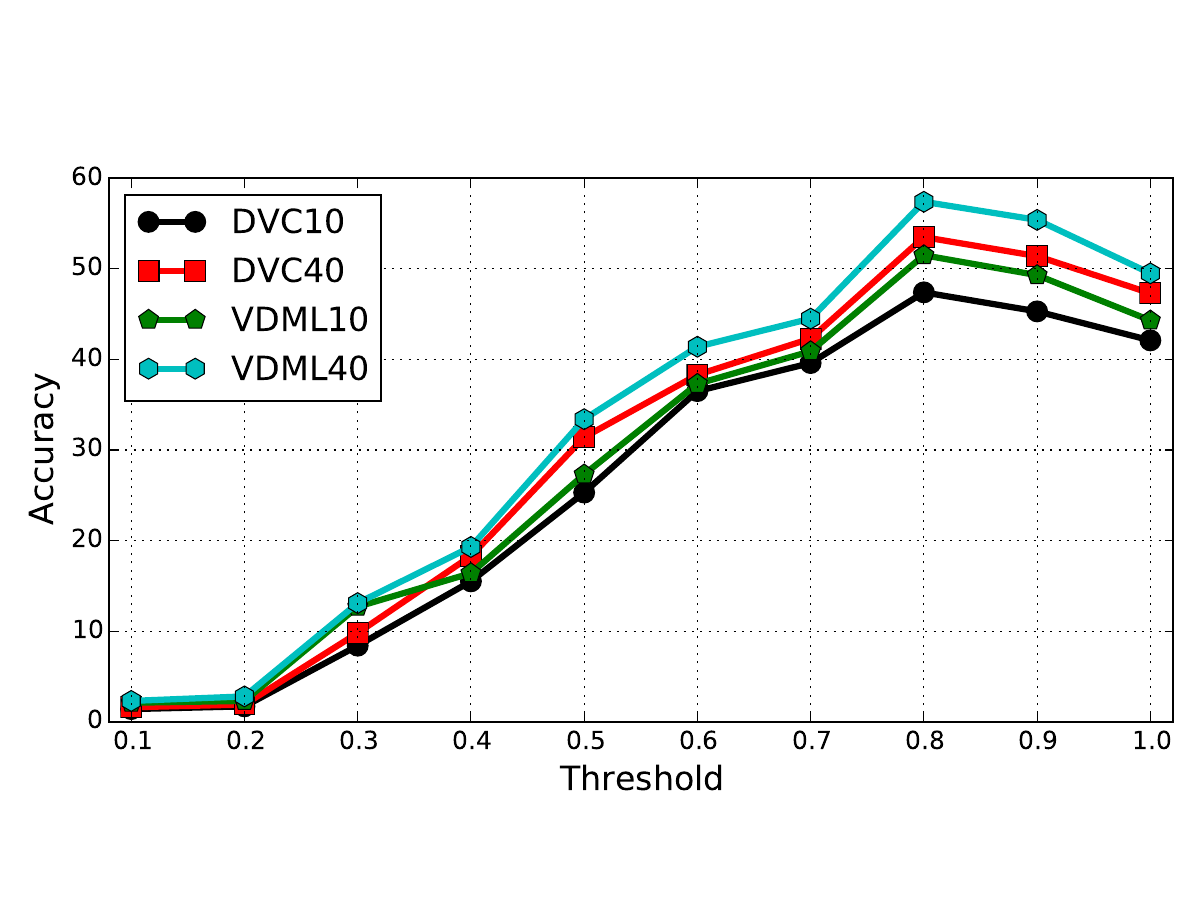}
    \vspace{-1em}
	\caption{Accuracy \vs threshold of supervoxel overlap with ground-truth for training on NYU
		Depth v2 dataset.
	}
	\label{fig:accvsth}
    \vspace{-1em}
\end{figure}

\noindent \vspace{-0.2em} \ \\
\textbf{Impact of ground-truth overlap threshold.} Fig.~\ref{fig:accvsth} shows the variation in
accuracy \vs threshold ($\beta$), for percentage of points that overlap with ground-truth, to consider it as a
positive supervoxel for training. The graph was obtained by splitting the train dataset into train
and validation sets with a ratio of $80$\% to $20$\% respectively and performing $5$-fold
cross-validation. It can be observed that the accuracy reaches a peak at $0.8$ for both DVC and VDML
trained on either of ModelNet10 or ModelNet40. An explanation of better performance of soft
supervoxel assignment is because the generated supervoxels do not always lie perfectly within object
boundaries. A hard assignment would therefore result in rejection of many supervoxels near the
object boundaries. Since our method maximizes the distance between supervoxels on boundaries and
neighbouring supervoxels (on background or other objects), a hard assignment has a negative impact
on the overall performance which is also validated by a drop of $7.2$\% and $7.9$\%
(hard-assignment) in accuracy as compared to soft-assignment ($\beta$ = $0.8$) on validation set
with VDML10 and VDML40 respectively. On the other hand, despite having lower validation set accuracy (and subsequently
on test set also) DVC witnesses a lower drop of $5.3$\% (DVC10) and $6.2$\%
(DVC40) as compared to VDML. This demonstrates that it is relatively less prone to supervoxel assignment
criteria \cf the proposed VDML. 

\begin{figure}
	\centering
	
	\begin{tabular}{ccc}
		\includegraphics[width=24mm]{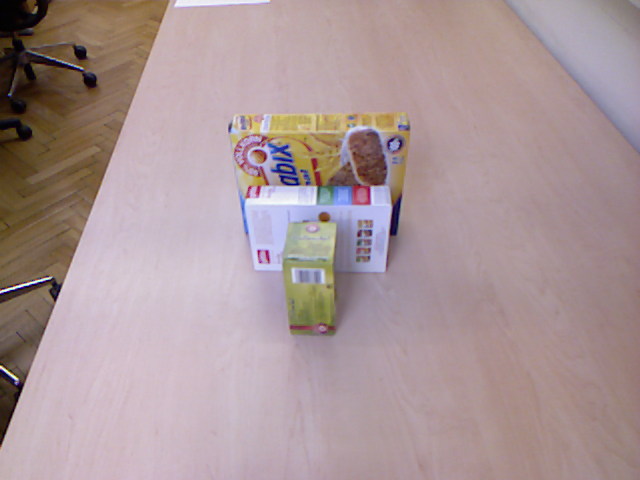} &   
		\includegraphics[width=24mm]{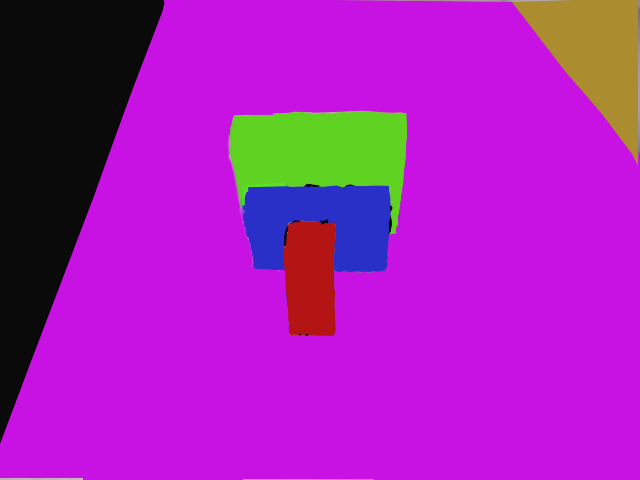} &
		\includegraphics[width=24mm]{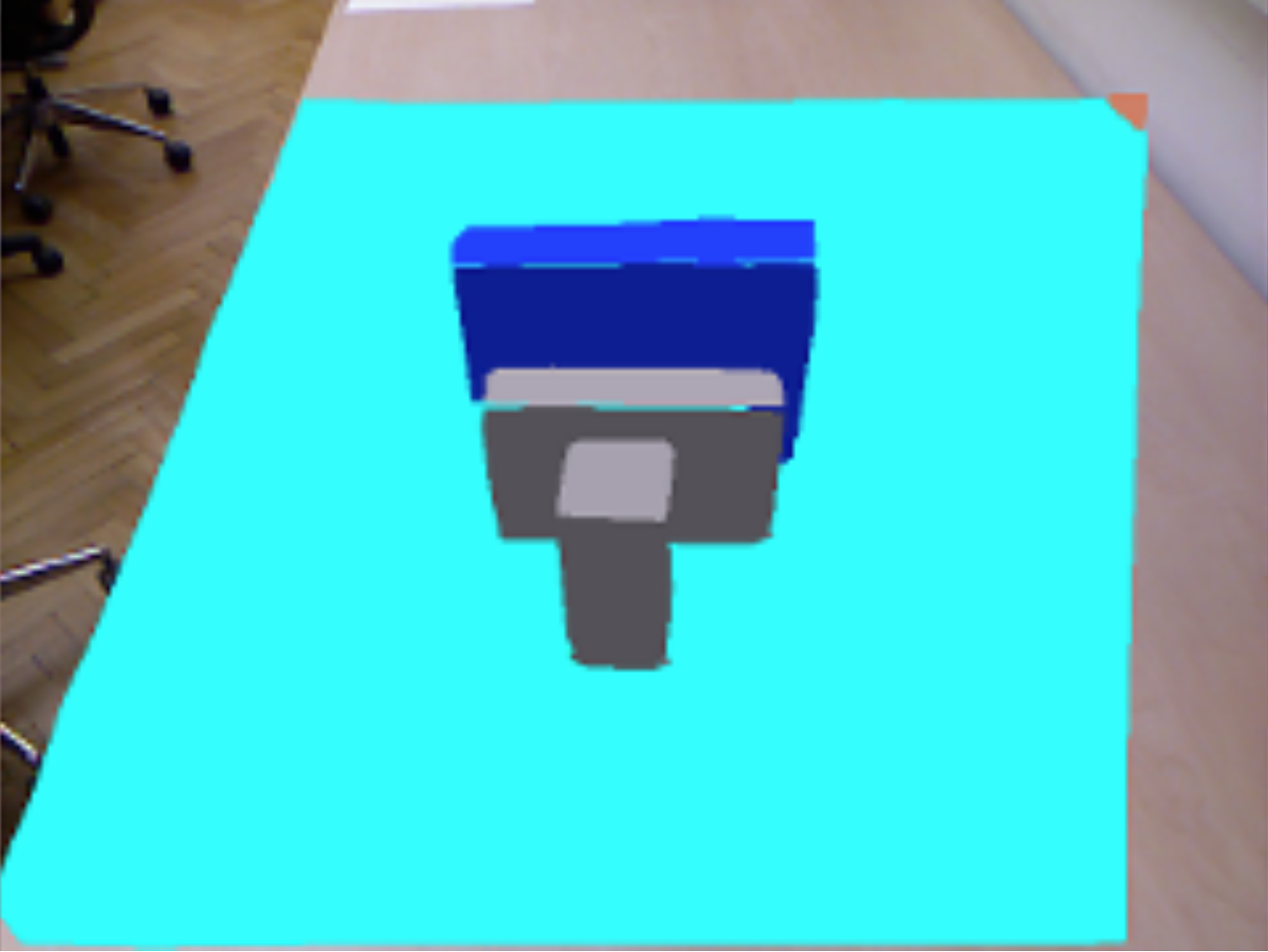} \\
		\includegraphics[width=24mm]{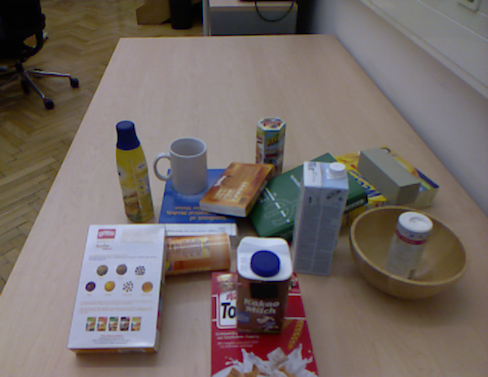} &  
		\includegraphics[width=24mm]{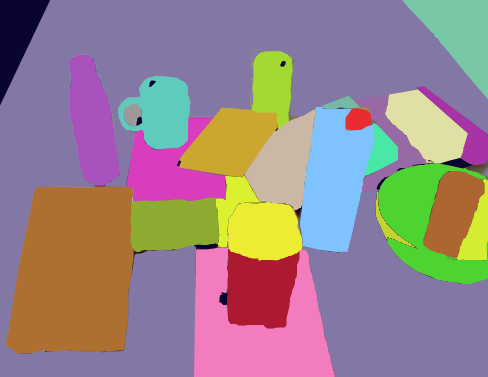} &
		\includegraphics[width=24mm]{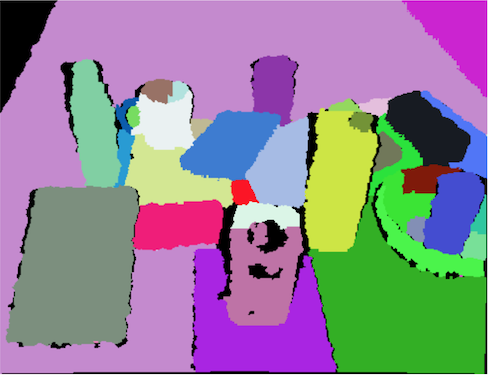}  \\
	\end{tabular}
    \caption{Qualitative results from Object Segmentation Dataset. First column shows the
    original images, second column shows the result using VDML40, while the third column shows
    results using \cite{richtsfeld2012segmentation} and \cite{christoph2014object} respectively.
    \vspace{2em}}
    \label{fig:osdcumu} \vspace{-1em}
\end{figure}

\begin{figure}
	\centering
	\scalebox{1}{
	\begin{tabular}{cc}
		\includegraphics[width=38mm]{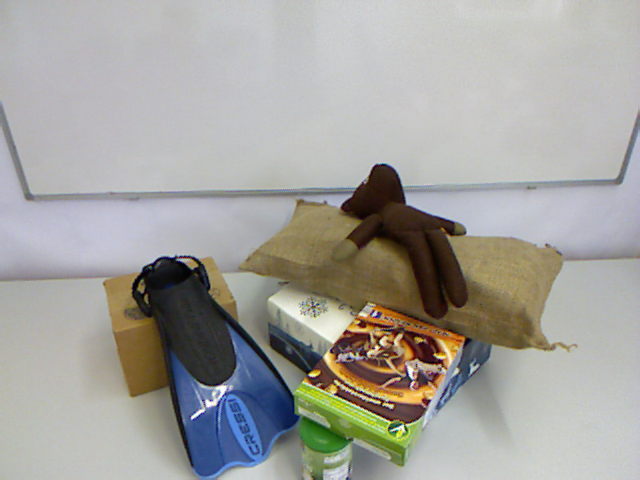} &   
		\includegraphics[width=38mm]{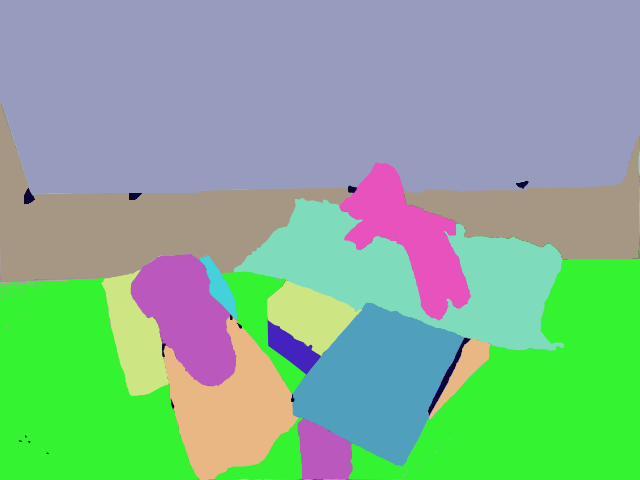} \\
		\includegraphics[width=38mm]{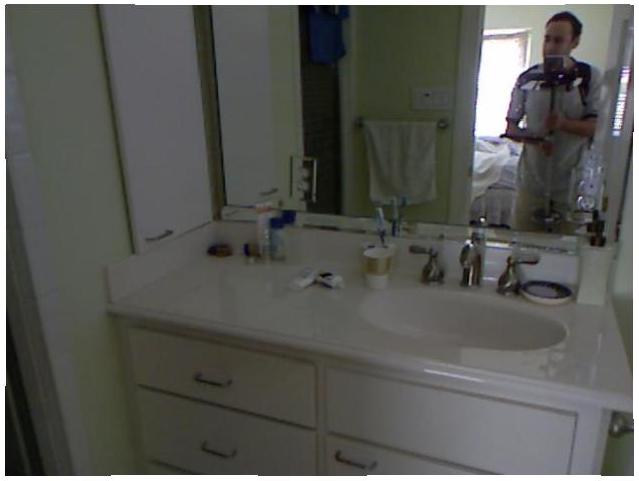} &  
		\includegraphics[width=38mm]{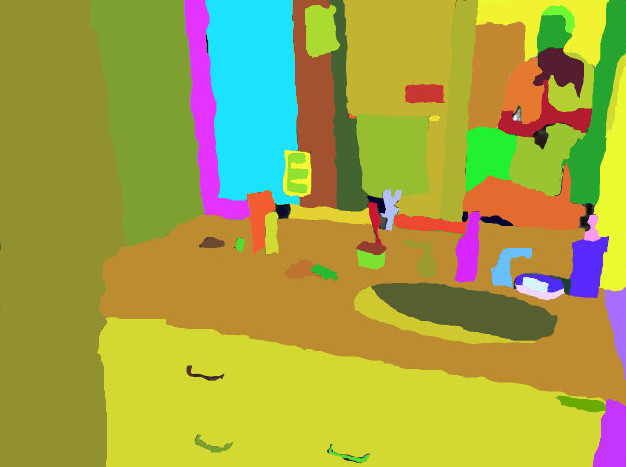} \\
	\end{tabular}}
    \caption{Left column has the original images while the right one shows the discovered objects
    using VDML40 from Object Discovery Dataset and NYU Depth v2 resp.
    }
    \label{fig:oddnyu} \vspace{-1em}
\end{figure}

\subsection{Qualitative Results}

Fig.~\ref{fig:osdcumu} shows visual comparison of results from Richtsfeld \etal \cite{richtsfeld2012segmentation} and  Stein \etal \cite{christoph2014object} on
Object Segmentation Dataset. In Fig.~\ref{fig:osdcumu} (c), the technique of
\cite{richtsfeld2012segmentation} tends to show under-segmentation and discovers two distinct
objects as a single object (grey). In comparison, VDML is able to clearly distinguish among
those objects. In Fig.~\ref{fig:osdcumu} (second row), we compare our approach on a scene with numerous occluded
objects. Fig.~\ref{fig:osdcumu} (f) shows the results provided by Stein \etal \cite{christoph2014object}.
On comparing Fig.~\ref{fig:osdcumu}(e) and \ref{fig:osdcumu}(f), it can be seen that VDML results
in lesser number of points classified as clutter (shown in black) while providing smoother
boundaries which would be helpful for applications such as robot grasping. In both the cases, the
can at the front suffers over-segmentation. In the first case, there are two visually arbitrary
segments while in the case of the proposed VDML those segments correspond to the body of the can and
top of the can. Moreover, the segments from the proposed VDML are closer to the ground-truth as
less number of points are marked as clutter.

Fig.~\ref{fig:oddnyu} (first row) shows results of our segmentation on Object Discovery Dataset.
There are two important observations to make here. First, as above, the boundaries are closer to the
ground-truth. This shows that VDML has learned to distinguish between object and non-object
boundaries since the training procedure explicitly focussed on supervoxel pairs around the object
boundaries. Secondly, we observe over-segmentation in the object towards the left in Fig
\ref{fig:oddnyu} (b) --the object gets divided into two segments. Although, we note that the two segments
actually belong to visually dissimilar parts of an object, which can easily be considered as
distinct objects in the absence of any context.

Fig.~\ref{fig:oddnyu} (second row)  shows results on NYU Depth v2 dataset. Since the previous visual images
consisted of objects of relatively similar sizes, we now demonstrate the results on scenes with
larger variation in sizes of objects. For example, the demonstrated image consists of objects like
a soap as well as a human. It can be observed that VDML is able to clearly distinguish among the objects in front of the mirror (Fig.~\ref{fig:oddnyu} (d)) except the
bottles, where we observe over-segmentation. The wash-basin also gets over-segmented, possibly
due to variation in depth and illumination across its surface. Although the objects in the mirror are difficult to discover due to lack of variation in depth, VDML still obtains distinguishable boundaries among various objects (towel, curtain) while over-segmenting the person holding the camera.

\section{Conclusion}
We proposed a novel object discovery algorithm which operates in the challenging learning based
setting where the objects to be discovered at test time were never seen at train time. We used
recent advances in 3D Convolutional Neural Networks and built on it a Siamese deep network for
learning non-linear embeddings of supervoxels into a Euclidean space which reflects object based
semantics. In the embeddings space the supervoxels from the same objects were constrained to be
closer than the supervoxels from different objects. Once these embeddings are learned they can be
used with supervoxels of different novel classes for doing object discovery using efficient
clustering in the embedding space.  We used a hybrid approach of using CAD models and subsequently
RGB-D images for learning non-linear embeddings of supervoxels for discovering objects that were never
seen before. We provided comprehensive empirical evaluation and comparison with several baselines
and existing methods to demonstrate the effectiveness of the technique. We showed that the
proposed architecture achieves high accuracy with low over and under segmentation rates making it
more reliable in the context of competing methods. We also demonstrated qualitatively the improved
performance of the proposed method.


{\small
	\bibliographystyle{ieee}
	\bibliography{IEEEfull}
}
	
	%
	
	
	
	
	
	
	

\end{document}